\documentclass[conference]{IEEEtran}
\IEEEoverridecommandlockouts

\usepackage{cite}
\usepackage{amsmath,amssymb,amsfonts}
\usepackage{algorithmic}
\usepackage{graphicx}
\usepackage{textcomp}
\usepackage{xcolor}
\usepackage{multirow}
\usepackage{colortbl}  
\usepackage{array}

\usepackage{stfloats}
\usepackage{url}
\usepackage{verbatim}
\usepackage{cite}
\usepackage{color}
\usepackage{booktabs}
\usepackage{utfsym}

\def\BibTeX{{\rm B\kern-.05em{\sc i\kern-.025em b}\kern-.08em
    T\kern-.1667em\lower.7ex\hbox{E}\kern-.125emX}}
\begin{document}

\title{MINR: Efficient \textbf{I}mplicit \textbf{N}eural \textbf{R}epresentations \\ for \textbf{M}ulti-Image Encoding\\

\thanks{*: Equal contributions. \textsuperscript{\dag}: Corresponding authors: Zhengwu Liu and Ngai Wong\{zwliu, nwong@eee.hku.hk\}. This work was supported in part by the Theme-based Research Scheme (TRS) project T45-701/22-R, National Natural Science Foundation of China (62404187) and the General Research Fund (GRF) Project 17203224, of the Research Grants Council (RGC), Hong Kong SAR.}
}

\author{
\IEEEauthorblockN{Wenyong Zhou*}
\IEEEauthorblockA{\textit{Department of EEE} \\
\textit{The University of Hong Kong }\\
Hong Kong SAR} \\
\IEEEauthorblockN{Yuxin Cheng}
\IEEEauthorblockA{\textit{Department of EEE} \\
\textit{The University of Hong Kong}\\
Hong Kong SAR}
\and

\IEEEauthorblockN{Taiqiang Wu*}
\IEEEauthorblockA{\textit{Department of EEE} \\
\textit{The University of Hong Kong}\\
Hong Kong SAR} \\
\IEEEauthorblockN{Chen Zhang}
\IEEEauthorblockA{\textit{Department of EEE} \\
\textit{The University of Hong Kong}\\
Hong Kong SAR} 
\and
\IEEEauthorblockN{Zhengwu Liu\textsuperscript{\dag}}
\IEEEauthorblockA{\textit{Department of EEE} \\
\textit{The University of Hong Kong}\\
Hong Kong SAR} \\
\IEEEauthorblockN{Ngai Wong\textsuperscript{\dag}}
\IEEEauthorblockA{\textit{Department of EEE} \\
\textit{The University of Hong Kong}\\
Hong Kong SAR}
}

\maketitle

\begin{abstract}
Implicit Neural Representations (INRs) aim to parameterize discrete signals through implicit continuous functions. However, formulating each image with a separate neural network~(typically, a Multi-Layer Perceptron (MLP)) leads to computational and storage inefficiencies when encoding multi-images. To address this issue, we propose MINR, sharing specific layers to encode multi-image efficiently. We first compare the layer-wise weight distributions for several trained INRs and find that corresponding intermediate layers follow highly similar distribution patterns. Motivated by this, we share these intermediate layers across multiple images while preserving the input and output layers as input-specific. In addition, we design an extra novel projection layer for each image to capture its unique features. Experimental results on image reconstruction and super-resolution tasks demonstrate that MINR can save up to 60\% parameters while maintaining comparable performance. Particularly, MINR scales effectively to handle 100 images, maintaining an average peak signal-to-noise ratio (PSNR) of 34 dB. Further analysis of various backbones proves the robustness of the proposed MINR.
\end{abstract}


\section{Introduction}
Implicit Neural Representations (INRs) have emerged as a powerful alternative to traditional discrete representations for various data types, including images, audio, and 3D shapes~\cite{chen2023neurbf, xie2023diner, zhang2024boosting}. Unlike conventional grid-based representations, INRs employ neural networks to encode signals as continuous functions, mapping spatial or temporal coordinates to the corresponding signal values~\cite{li2024learning, saragadam2023wire}. Such a continuous nature of INR enables high-resolution representations that can be queried at arbitrary points, offering significant advantages in terms of scalability and flexibility~\cite{Tancik2020Fourier,Mescheder2019OccupancyNetworks}. Recent advancements have demonstrated the effectiveness of INRs in various applications, from novel view synthesis~\cite{mildenhall2020nerf} to 3D shape reconstruction~\cite{Park2019DeepSDF} and audio signal processing~\cite{Sitzmann2020SIREN}, highlighting their versatility across different domains.
 
\begin{figure}[!t]
\centering
\includegraphics[scale=0.55]{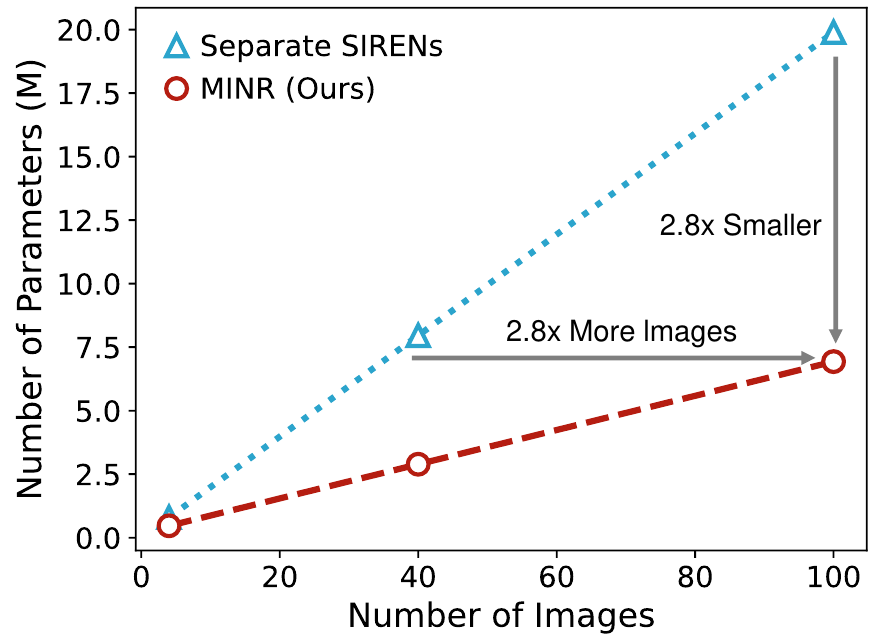}
\caption{Relation between the encoded images and parameters. MINR can save 2.8$\times$ parameters than separate SIRENs when encoding 100 images.} 
\label{fig:param_compar}
\vspace{-0.2cm}
\end{figure}

However, classical INR requires a separate neural network (typically, a \textit{Multi-Layer Perceptron (MLP)}) for each signal~\cite{mcginnis2023single, strumpler2022implicit, liu2024finer}. While such a design effectively captures high-fidelity representations, it also leads to significant parameter redundancy and storage inefficiencies, particularly when encoding multi-images~\cite{dupont2021coin, dupontcoin++}. Specifically, traditional INR encodes each image as an individual MLP and thus brings a rapid increase in storage memory for multiple images, making it impractical for many real-world applications~\cite{zhang2024ntinr, liasmr}.
As shown in Figure~\ref{fig:param_compar}, if we encode each image with one SIREN model (denoted as separate SIRENs~\cite{Sitzmann2020SIREN}), the parameters would increase rapidly.
Meanwhile, our proposed MINR would reduce the parameters to about one-third to encode 100 images.

\begin{figure*}[!t]
\centering
\includegraphics[scale=0.23]{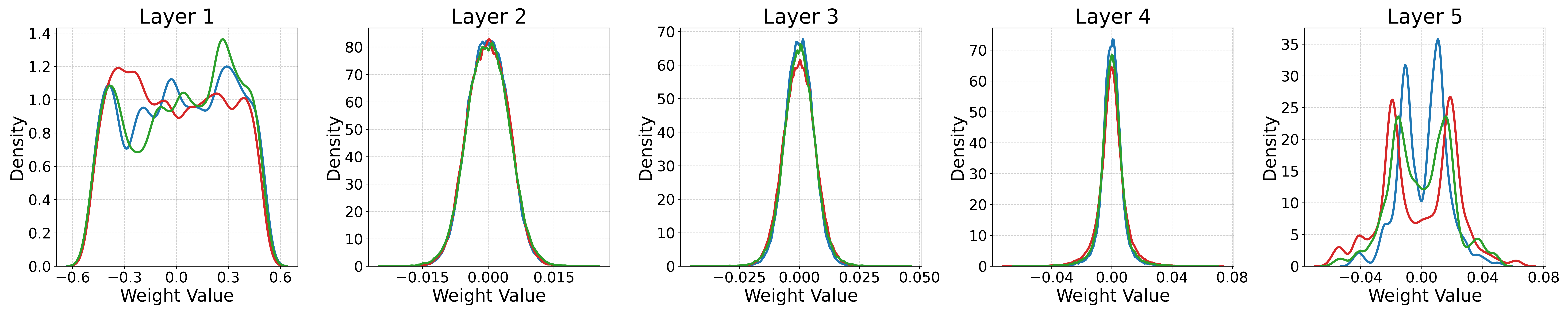}
\caption{The layer-wise weight distributions for different baseline models trained on {\tt Kodak01} (red), {\tt Kodak03} (green), and {\tt Kodak18} (blue) images, separately. The weights in the \textit{three intermediate} layers share \textbf{similar} distribution patterns, while the \textit{first} and \textit{last} layers are \textbf{input-specific}.}
\label{fig:motivation}
\vspace{-0.2cm}
\end{figure*}

To address this issue, we propose MINR, aiming to encode multi-images efficiently. We first compare the layer-wise weight distributions among different images and reveal that different INRs share the same patterns with respect to intermediate layers, as shown in Figure~\ref{fig:motivation}. This observation motivates us to share these intermediate layers across multiple images, which contain the majority of the parameters. Therefore, we can encode multi-images efficiently. In addition, a custom projection layer is introduced before the shared layers to capture more image-specific information. Experimental results on the image reconstruction and super-resolution tasks demonstrate that MINR can save up to 60\% parameters. Moreover, MINR can effectively scale to 100 images, showing its robustness. In summary, the main contributions of this paper are as follows:
\begin{itemize} 
    \item We explore the layer-wise weight distributions of MLPs across different images, revealing that the intermediate layers exhibit a similar distribution pattern while the first and last layers are image-specific.
    \item Based on the observation, we propose a novel MINR framework to encode multiple images efficiently by sharing the intermediate layers. Moreover, we introduce a customized projection layer to capture the specific information for each image. 
    \item We conduct extensive experiments on image reconstruction and super-resolution tasks. Experimental results demonstrate the effectiveness and robustness of MINR for multi-image encoding. 
\end{itemize}
\section{Methodology}
\label{sec:method}
In this section, we first present the layer-wise weight distributions of various INRs. The intermediate layers show similar patterns. Based on this observation, we then propose a novel MINR framework to encode multi-images efficiently.
\subsection{Layer-wise Weight Distributions}
An INR maps a coordinate $\mathbf{x} \in \mathbb{R}^d$ to a signal value $\mathbf{y} \in \mathbb{R}^k$. The function approximated by the INR can be described as $\mathbf{y}=f_\theta(\mathbf{x})$, where $f_{\theta}$ represents the model parameters~\cite{peng2021neural, saragadam2022miner, xu2022signal}. Among various INR models, the SIREN model, which utilizes sinusoidal activation functions, is particularly effective in modeling high-frequency details. Its ability to capture intricate patterns makes it a natural choice for our study. 

We reconstruct images in the Kodak dataset and analyze the weight distributions across all five layers of different models, which is shown in Figure~\ref{fig:motivation}. Although uniformly initialized, the weights in the intermediate layers tend to follow a bell distribution, while the weights in the first layer remain closer to the original uniform distribution, and those in the last layer resemble a bimodal distribution. A key observation is the \textit{diversity} in the weight distributions of the first and last layers compared to the \textit{similarity} in the weight distributions of the intermediate layers. 
The first layer in each model exhibits a diverse and relatively wide distribution of weights, highlighting its role in transforming input coordinates into a high-dimensional feature space. This transformation is crucial for capturing the unique spatial characteristics of each image, with the broad range of weights reflecting the need to represent a wide spectrum of frequencies effectively.
Similarly, the last layer also shows a diverse weight distribution. This is expected, as the last layer is responsible for mapping the high-dimensional representations back to the output space (e.g., RGB values for images). The variation in weight distribution across different images in this layer is necessary to adapt to the specific details of each image. In contrast, the similar Gaussian distribution observed in the intermediate layers suggests that these layers are primarily fine-tuning the representations learned in the first layer rather than introducing significant transformations. This uniformity across images indicates that the intermediate layers could be shared among different images with minimal impact on the model's performance.
\subsection{MINR for Multi-Image Encoding}
The observation that intermediate layers in different INRs exhibit similar weight distributions suggests the potential for parameter sharing. Specifically, the intermediate layers seem to perform more general feature extraction, which can be effectively shared across multiple images, while the first and last layers are more critical for handling the distinct details specific to each image. By sharing these intermediate layers across multiple images, we can effectively reduce the redundancy seen in traditional INR models, leading to significant savings in the total number of parameters required. 

Moreover, while sharing intermediate layers across multiple images can enhance efficiency, it may also reduce the model's ability to accurately capture the unique features of each image. To mitigate this potential issue, we introduce a projection layer before the shared intermediate layers. This projection layer allows the model to incorporate the unique characteristics of each image from the beginning of the reconstruction pipeline, ensuring the model does not lose its capacity to differentiate between images. Therefore, MINR achieves a trade-off between efficiency and effectiveness.

\begin{figure}[!t]
\centering
\includegraphics[scale=0.25]{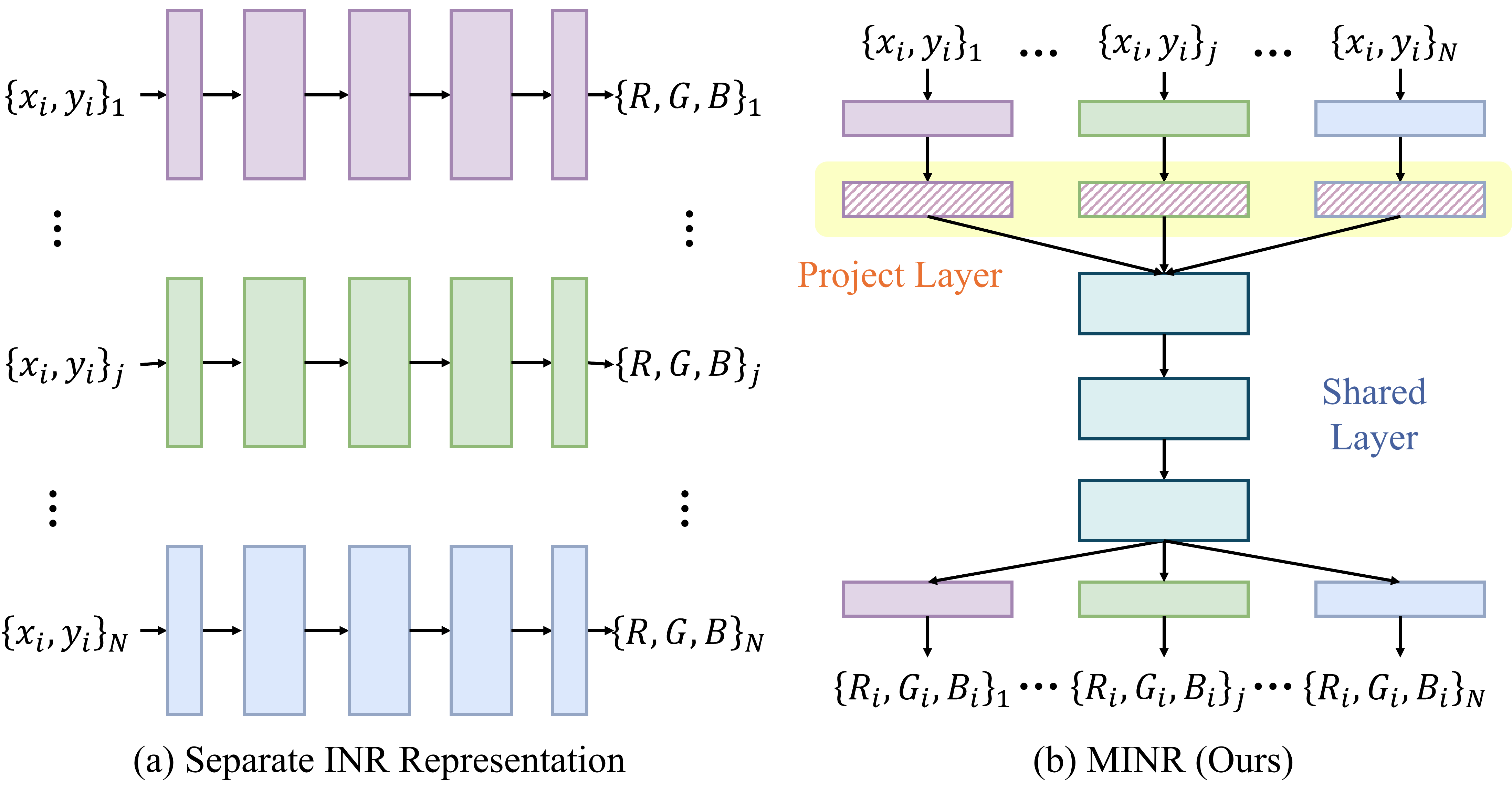}
\caption{The framework of separate SIRENs (part a) and MINR (part b). 
Each image has input and output layers, an extra projection layer, and \textit{shared} intermediate layers.} 
\label{fig:framework}
\vspace{-0.2cm}
\end{figure}

Figure~\ref{fig:framework} shows the separate SIRENs (part a) and an overview of our proposed MINR (part b). For the separate SIRENs, each image is encoded by one SIREN model. In MINR, the first layer remains image-specific, transforming the input coordinates $\mathbf{x}$ for each image $i$ into a hidden feature vector $\mathbf{z}_{in}^{(i)}$:
\begin{equation} 
\mathbf{z}_{in}^{(i)}=\sigma(\mathbf{W}_{in}^{(i)}\mathbf{x} + \mathbf{b}_{in}^{(i)})
\end{equation}
where $\mathbf{W}_{in}^{(i)}$ and $\mathbf{b}_{in}^{(i)}$ are the image-specific weight and bias parameters and $\sigma(x)$ is the nonlinear activation function. After this initial transformation, we insert an additional image-specific layer $f_{proj}^{(i)}$, which further processes $\mathbf{z}_{in}^{(i)}$ before it is passed through the shared intermediate layers:
\begin{equation} 
\mathbf{z}_{proj}^{(i)}=f_{proj}^{(i)}(\mathbf{z}_{in}^{(i)})
\end{equation}
Followed by the shared intermediate layer:
\begin{equation} 
\mathbf{z}_{l+1}=\sigma(\mathbf{W}_l \mathbf{z}_l+\mathbf{b}_l) \quad l = 1, 2, 3
\end{equation}
where $\mathbf{W}_{l}$ and $\mathbf{b}_{l}$ are the weights and biases of the shared layer $l$, $\mathbf{z}_{l}$ is the input to layer $l$, and $z_1=z_{in}$. 
These shared layers refine the image-specific features into a latent representation that captures the general patterns common to all images in the dataset. 
Finally, the output of the shared layers is passed through an image-specific output layer $f_{out}^{(i)}$:
\begin{equation} 
\mathbf{y}^{(i)}= f_{out}^{(i)}(\mathbf{z}_{3})=\mathbf{W}_{out}^{(i)}\mathbf{z}_{3}+\mathbf{b}_{out}^{(i)}
\end{equation}

During inference, MINR reconstructs images by taking the input image size and dynamically loading the corresponding image-specific layers to construct the model. The reconstruction pipeline remains consistent, and this approach significantly reduces memory storage requirements while maintaining the model's flexibility and efficiency.

\section{Experiments}
In this section, we compare MINR with several previous methods to encode multiple images on the image reconstruction and super-resolution task.
\subsection{Experimental Setup} 
We employ the SIREN as the baseline model~\cite{Sitzmann2020SIREN} and verify the robustness of MINR across various INRs. All experiments are conducted on a GeForce RTX 3090 GPU.

\subsection{Experimental Results}
\begin{table}[!t]
\centering
\renewcommand{\arraystretch}{1.2}
\caption{Comparison between our method and the original SIREN model by individual training and with different image concatenation (including concatenation in row, column, grid, or along the third ID dimension).}
\definecolor{gg}{HTML}{e2f0cb}
\resizebox{\linewidth}{!}{
\begin{tabular}{lccc}
\toprule
\textbf{Methods} & \textbf{Params (M)} & \textbf{PSNR (dB) $\uparrow$ } & \textbf{SSIM $\uparrow$} \\
\midrule
\multicolumn{4}{c}{\textit{4-Image Reconstruction (Kodak)}} \\
\midrule
Separate SIRENs & 0.80 & 38.24\textsubscript{$\pm$0.58} & 0.96\textsubscript{$\pm$0.01} \\
SIREN (Row)                       & 0.20                              & 30.19\textsubscript{$\pm$0.67}  & 0.83\textsubscript{$\pm$0.05} \\
SIREN (Column)                     & 0.20                              & 31.05\textsubscript{$\pm$0.43}     & 0.84\textsubscript{$\pm$0.06} \\
SIREN (Grid)                       & 0.20                              & 30.96\textsubscript{$\pm$ 0.96}     & 0.83\textsubscript{$\pm$0.05} \\
SIREN (ID)                      & 0.20                              & 31.54\textsubscript{$\pm$0.69}      & 0.85\textsubscript{$\pm$0.03} \\
COIN++~\cite{dupontcoin++}                      & 0.20                              & 31.27\textsubscript{$\pm$0.82}      & 0.85\textsubscript{$\pm$0.03} \\
\rowcolor{gg} MINR (Ours) & 0.47 & 36.24\textsubscript{$\pm$0.68} & 0.93\textsubscript{$\pm$0.01} \\
\midrule
\multicolumn{4}{c}{\textit{40-Image Reconstruction (ImageNet)}} \\
\midrule
Separate SIRENs & 7.95 & 38.17\textsubscript{$\pm$1.42} & 0.96\textsubscript{$\pm$0.01} \\
SIREN (Row)                        & 0.20                              & 28.16\textsubscript{$\pm$1.58}   & 0.80\textsubscript{$\pm$0.06} \\
SIREN (Column)                     & 0.20                              & 28.24\textsubscript{$\pm$1.64}     & 0.80\textsubscript{$\pm$0.06} \\
SIREN (Grid)                       & 0.20                              & 28.53\textsubscript{$\pm$1.49}     & 0.80\textsubscript{$\pm$0.06} \\
SIREN (ID)                      & 0.20                              & 29.16\textsubscript{$\pm$1.57}       & 0.80\textsubscript{$\pm$0.03} \\
COIN++~\cite{dupontcoin++}                      & 0.33                              & 30.42\textsubscript{$\pm$1.73}      & 0.87\textsubscript{$\pm$0.05} \\
\rowcolor{gg} MINR (Ours) & 2.90 & 34.37\textsubscript{$\pm$1.29} & 0.91\textsubscript{$\pm$0.01} \\
\midrule
\multicolumn{4}{c}{\textit{100-Image Reconstruction (ImageNet)}} \\
\midrule
Separate SIRENs & 19.88 & 38.06\textsubscript{$\pm$1.58} & 0.96\textsubscript{$\pm$0.01} \\
\rowcolor{gg} MINR (Ours) & 6.93 & 34.02\textsubscript{$\pm$1.67} & 0.93\textsubscript{$\pm$0.01} \\
\midrule
\multicolumn{4}{c}{\textit{Video}} \\
\midrule
Separate SIRENs & 9.94 & 38.95\textsubscript{$\pm$0.24} & 0.96\textsubscript{$\pm$0.01} \\
\rowcolor{gg} MINR (Ours) & 3.56 & 35.93\textsubscript{$\pm$1.58} & 0.94\textsubscript{$\pm$0.01} \\
\bottomrule
\end{tabular}
}
\label{tab:result}
\end{table}

\noindent \textbf{Baselines.}
Besides the individual INR for each image (Separate SIRENs) and COIN++ \cite{dupontcoin++}, we also introduce more baselines. Specifically, we concatenate the images into one larger image (by row, column, and grid) and encode it using one SIREN model. In addition, we can consider multiple images as 3D data and add an ID index to the coordinates, which is denoted as SIREN (ID).

\noindent \textbf{Results on Image Reconstruction.} 
Table \ref{tab:result} shows the results of the image reconstruction task. When encoding 4 images, MINR reduces the number of parameters by 41.25\% compared to separate SIRENs, while achieving comparable performance. To explore the capability to encode more images, we randomly sample some images from the ImageNet \cite{deng2009imagenet} training set. For 40 images, MINR continues to maintain an average PSNR value above 34 dB, while compressing the parameters by 63.53\% compared to the separate approach. Even for 100 images, MINR performs well, retaining an average PSNR above 34 dB with only 34.85\% of the parameters compared to separate SIRENs. Moreover, for a video with 50 frames, MINR gets a PSNR of 35.93. The findings are similar considering the SSIM metric. MINR outperforms all the baselines adapting one SIREN model. In contrast, the image concatenation baselines underperform even with just four images, primarily due to the reduced resolution resulting from concatenation. The performance gap between image concatenation methods widens as the number of images increases.
\begin{figure}[!t]
\centering
\includegraphics[scale=0.29]{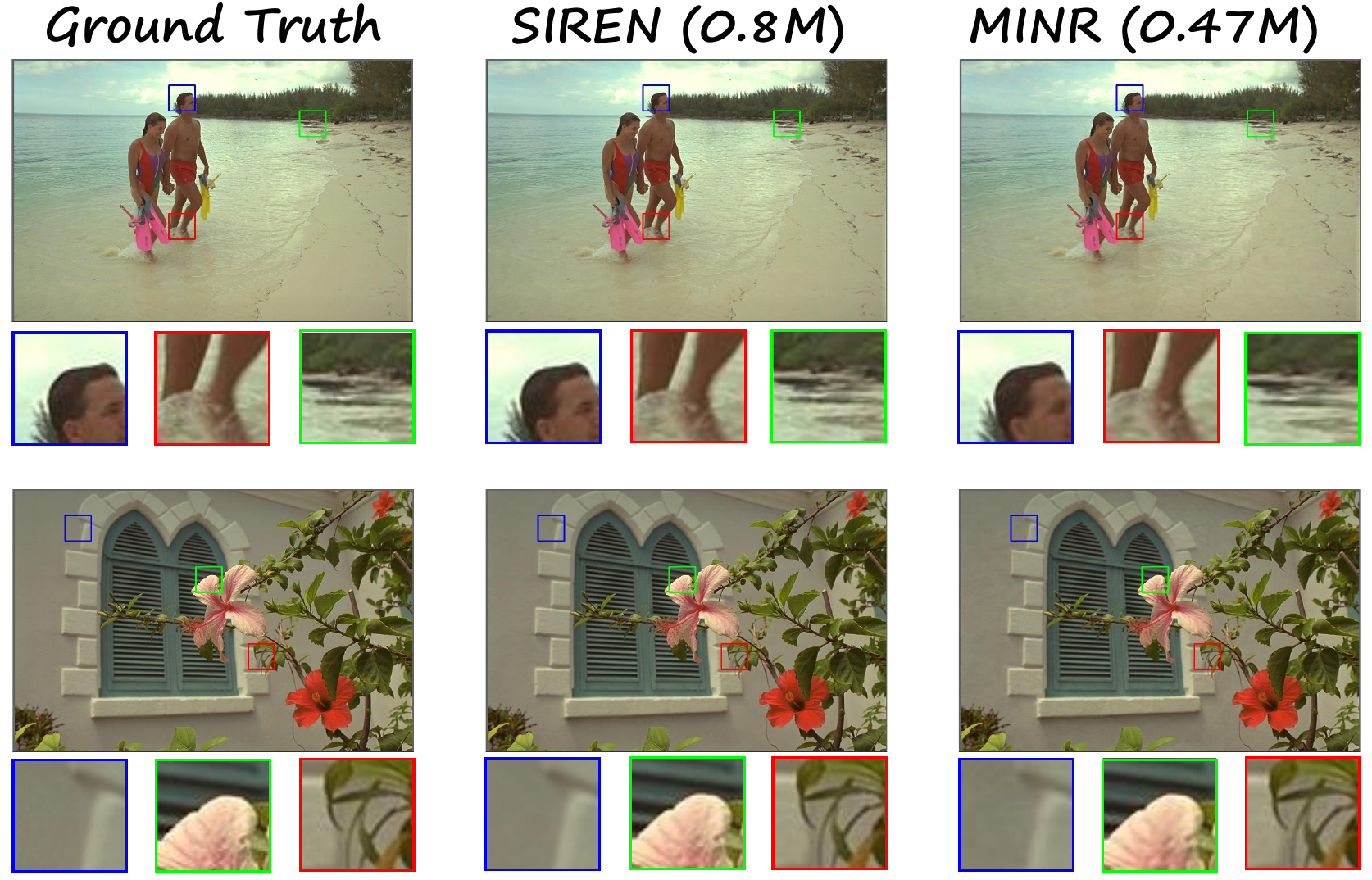}
\caption{Visualization of the image \textbf{super-resolution} results including ground truth, SIREN, and MINR on {\tt Kodak07} and {\tt Kodak12} images. Both 2$\times$ (1st row) and 4$\times$ (2nd row) super-resolution are reported.} 
\label{fig:visual_result2}
\vspace{-0.2cm}
\end{figure}
\begin{table}[!t]
\centering
\caption{Comparison between separate INRs and our MINR method on 4-image encoding on the Kodak dataset.}
\renewcommand{\arraystretch}{1.2}
{
\resizebox{0.85\linewidth}{!}{
\begin{tabular}{lcc}
\toprule
\textbf{Methods}          & \textbf{Params (M)} & \textbf{PSNR (dB) $\uparrow$} \\ \hline
Separate \textbf{Gausses}~\cite{ramasinghe2022beyond}  & 0.80         & 38.13           \\
\quad MINR (Ours)                   & 0.47         & 35.94        \\ 
 Separate \textbf{WIREs}~\cite{saragadam2023wire}  & 0.80         & 38.24           \\
\quad MINR (Ours)                   & 0.47         & 36.61        \\  
Separate \textbf{FINERs}~\cite{liu2024finer}  & 0.80         & 38.92           \\
\quad MINR (Ours)                   & 0.47         & 36.87        \\
\bottomrule
\end{tabular}}
}
\label{tab:backbone}
\vspace{-0.2cm}
\end{table}

\noindent \textbf{Results on Image Super-resolution.}
Moreover, we evaluated our method on super-resolution tasks, with visualization results shown in Figure \ref{fig:visual_result2}. 
We trained the models using images from the Kodak dataset, downsampled by factors of 2 and 4, and then reconstructed them to their original size. Compared to the separate representations, which achieved 38.87 dB and 37.25 dB on two images, our method reduces the number of parameters by over 40\% while maintaining similar results with PSNR values of 36.59 dB and 35.91 dB. For a better comparison, we selected three regions in each image that are typically prone to losing detail during upscaling due to the large density variation. The amplified regions highlight the effectiveness of our approach in preserving high-quality reconstructions while significantly reducing computational costs.
\subsection{Ablation Study}
\noindent \textbf{Ablation on Various Backbones.} Table~\ref{tab:backbone} compares separate representations and our method for 4-image encoding across different INR models, including Gaussian-based INR \cite{ramasinghe2022beyond}, WIRE \cite{saragadam2023wire}, and FINER \cite{liu2024finer}. These models introduce various activation functions to address the limitations of SIREN, so the number of parameters remains the same across them. 
Compared to separate representations, our proposed MINR consistently demonstrates comparable performance for multi-image encoding while offering significant parameter reduction, showing the robustness of MINR.

\noindent \textbf{Ablation on Shared Layers.} 
Table~\ref{tab:ablation} presents another ablation study in our design. Using MINR as the baseline, we first remove the image-specific projection layers while keeping all other configurations identical. This modification results in a significant degradation of over 9 dB in PSNR, highlighting the crucial role of the projection layers in providing sufficient parameters to learn image-specific features. 
Next, we remove one intermediate layer from the shared layers to introduce more parameters for extracting image-specific information. Although this change slightly improves the PSNR, the disproportionate increase in model size indicates that this approach is not cost-effective. Additionally, we observe that removing layers closer to the input from the shared layers yields better performance, further supporting the importance of image-specific projection layers. Finally, we remove two layers from sharing layers, which increases the number of parameters but only marginally improves performance. This outcome suggests that our original design effectively balances hardware efficiency and model performance. In conclusion, our proposed MINR can achieve a good trade-off between effectiveness and efficiency.

\begin{table}[!t]
\centering
\caption{The ablation study of freezing different layers in the SIREN model on a 4-image reconstruction task on the Kodak dataset.}
\renewcommand{\arraystretch}{1.2}
{
\resizebox{\linewidth}{!}{
\begin{tabular}{lcc}
\toprule
\textbf{Methods}          & \textbf{Params (M)} & \textbf{PSNR (dB) $\uparrow$} \\ \hline
\textbf{MINR (Ours)}  & 0.47         & 36.24           \\
\quad \textit{w}/\textit{o} project layers                   & 0.27         & 27.15        \\  
\quad \textit{w}/\textit{o} sharing 2nd layer  & 0.66         & 36.53           \\
\quad \textit{w}/\textit{o} sharing 3rd layer  & 0.66         & 36.49          \\
\quad \textit{w}/\textit{o} sharing 4th layer  & 0.66         & 36.39           \\
\quad \textit{w}/\textit{o} sharing 2nd \& 3rd layers  & 0.86         & 36.67           \\
\quad \textit{w}/\textit{o} sharing 2nd \& 4th layers  & 0.86         & 36.75           \\
\quad \textit{w}/\textit{o} sharing 3rd \& 4th layers  & 0.86         & 36.62           \\
\bottomrule
\end{tabular}}
}

\label{tab:ablation}
\vspace{-0.2cm}
\end{table}

\section{Conclusion}
\label{sec:conclusion}
This paper presents an efficient approach to encoding multiple images, overcoming the limitations of classical INRs that require separate training and storage. By leveraging the observed similarity in weight distributions across intermediate layers in different INRs, we propose a novel MINR that shares these layers among multiple images to capture common features. To retain the unique information of each image, we keep the input and output layers image-specific and introduce an extra projection layer before the shared layers. Experimental results show that our approach reduces model parameters by up to 60\% compared to separate training strategies while maintaining representation quality above 34 dB per image. Further analysis indicates that MINR achieves a trade-off between effectiveness and efficiency. 

\vfill\pagebreak

\label{sec:refs}
\bibliographystyle{IEEEbib}
\bibliography{refs}

\end{document}